\documentclass[sigconf]{acmart}
\usepackage{enumitem}
\usepackage{balance} 
\AtBeginDocument{%
  \providecommand\BibTeX{{%
    \normalfont B\kern-0.5em{\scshape i\kern-0.25em b}\kern-0.8em\TeX}}}

\copyrightyear{2022} 
\acmYear{2022} 
\setcopyright{acmcopyright}\acmConference[CIKM '22]{Proceedings of the 31st ACM International Conference on Information and Knowledge Management}{October 17--21, 2022}{Atlanta, GA, USA}
\acmBooktitle{Proceedings of the 31st ACM International Conference on Information and Knowledge Management (CIKM '22), October 17--21, 2022, Atlanta, GA, USA}
\acmPrice{15.00}
\acmDOI{10.1145/3511808.3557224}
\acmISBN{978-1-4503-9236-5/22/10}
\renewcommand{\shortauthors}{Wei Shen et al.}

\begin{document}


\title{A Transformer-Based User Satisfaction Prediction for Proactive Interaction Mechanism in DuerOS
}

\author{Wei Shen}
\email{shenwei07@baidu.com}
\affiliation{
  \institution{Baidu Inc.}
  \city{Beijing}
  \country{China}
}

\author{Xiaonan He}
\authornote{Both authors are corresponding authors.}
\email{hexiaonan@baidu.com}
\affiliation{
  \institution{Baidu Inc.}
  \city{Beijing}
  \country{China}
}

\author{Chuheng Zhang}
\email{zhangchuheng123@live.com}
\affiliation{
  \institution{IIIS, Tsinghua University}
  \city{Beijing}
  \country{China}
}

\author{Xuyun Zhang}
\email{xuyun.zhang@mq.edu.au}
\affiliation{
  \institution{Macquarie University}
  \city{Sydney}
  \country{Australia}
}

\author{Jian Xie}
\authornotemark[1]
\email{xiejian01@baidu.com}
\affiliation{
  \institution{Baidu Inc.}
  \city{Beijing}
  \country{China}
}

\begin{abstract}
Recently, spoken dialogue systems have been widely deployed in a variety of applications, serving a huge number of end-users.
A common issue is that the \emph{errors} resulting from noisy utterances, semantic misunderstandings, or lack of knowledge make it hard for a real system to respond properly, possibly leading to an unsatisfactory user experience.
To avoid such a case, we consider a proactive interaction mechanism where the system predicts the user satisfaction with the candidate response before giving it to the user. 
If the user is not likely to be satisfied according to the prediction, the system will ask the user a suitable question to determine the real intent of the user instead of providing the response directly. 
With such an interaction with the user, the system can give a better response to the user. 
Previous models that predict the user satisfaction are not applicable to DuerOS which is a large-scale commercial dialogue system.
They are based on hand-crafted features and thus can hardly learn the complex patterns lying behind millions of conversations and temporal dependency in multiple turns of the conversation. 
Moreover, they are trained and evaluated on the benchmark datasets with adequate labels, which are expensive to obtain in a commercial dialogue system. 
To face these challenges, we propose a pipeline to predict the user satisfaction to help DuerOS decide whether to ask for clarification in each turn. 
Specifically, we propose to first generate a large number of weak labels and then train a transformer-based model to predict the user satisfaction with these weak labels. 
Moreover, we propose a metric, contextual user satisfaction, to evaluate the experience under the proactive interaction mechanism.  
At last, we deploy and evaluate our model on DuerOS, and observe a 19\% relative improvement on the accuracy of user satisfaction prediction and 2.3\% relative improvement on user experience.
\end{abstract}

\ccsdesc[500]{Computing methodologies~Discourse, dialogue and pragmatics}

\keywords{Dialogue System, Transformer-based Model, User Satisfaction Prediction}

\maketitle
\section{Introduction}
DuerOS from Baidu Inc. is a commercial spoken dialogue system that serves millions of users, through conversations, to complete a series of tasks such as acquiring weather forecasts, such as searching for flight information and playing music. Like Alexa, Google Assistant, and other modern conversational systems, DuerOS processes a user utterance and responses to the user by a pipeline of the following modules: 
an automatic speech recognition (ASR) module \cite{huang2019study, golan2018deep}, a language understanding (LU) module \cite{zhang2019joint}, 
a knowledge graph (KG) module \cite{shen2019multi-task, igor2019data-effi}, 
an information retrieval (IR) module \cite{kasemsap2017mastering}, a natural language generation (NLG) module \cite{buck2018ask, li2017end, wen2016network},
a dialogue manager (DM) module \cite{nikola2017neural, kuma2019practical, chen2019seman} and a text-to-speech (TTS) module \cite{peng2019parallel}.
However, the users interacting with DuserOS voice assistant may experience frictions due to various reasons: 1) Automatic Speech Recognition (ASR) errors, such as misrecognizing the user utterance as some unrecognizable tokens, 2) Natural Language Understanding (NLU) errors, such as misunderstanding the user utterance that aims to play a song as intending to play a short video, 3) and user errors, such as speaking of playing a song as stopping playing a song. To ensure the success of a conversational system, it is essential to fix these frictions to let users have a more seamless and engaged experience.

Existing commercial spoken dialogue systems usually adopt an error correction module to \emph{fix} these errors, such as the query reformulation module in DuerOS and Alexa \cite{ponnusamy2020feedback}. In the query reformulation module, the system directly reformulates the user utterance if an error is detected in the user utterance. However, the error correction module can only avoid a small fraction of user frictions. (Specially, it can fix only around 10\% of the user frictions in DuerOS.) The remaining frictions in the system can still make the users hard to enjoy a seamless experience. Moreover, the correction may be wrong and further confuses the user.

%Fig. \ref{fig:Intro} shows an example where a user gives the utterance \say{Hey Xiaodu, play the song \emph{show up}} to DuerOS, while the real intent is \say{play the song \emph{show me love}}. Based on the user's historical utterances, the error correction module can correct the name of the song. This correction prevents the system from giving a wrong response to the user, which improves the user experience to a certain extent. However, the error correction module can only avoid a small fraction (about 10 percent) of user frictions in DuerOS. The remaining frictions in the system make users hard to enjoy a seamless experience. Moreover, the correction may be wrong and confuses the user. In the above example, if the user actually hopes the system to play the song \emph{show up}, correcting the user query and play the song \emph{show me love} would lead to new user frictions instead. 
A better way to address these frictions is to proactively interact with the user (i.e., ask the user a clarification question and get his/her answer to this question) when the system recognizes that the candidate response might fail to satisfy the user. With the interaction with the user, the system can prepare a better response. Specifically, the system first uses a predictor to predict the user satisfaction with the candidate response. According to the user satisfaction, the system decides whether to ask for clarification or directly give this candidate response. Instead of correcting the user utterance directly, asking for clarification is a better way for the system to interact with the user when the system cannot determine the real intent of the user. In addition, some users in DuerOS tend to modify a previous utterance in hopes of fixing an error in the previous turn (which may be an error from the user or the system) to get the right results, but some users may give up or change to another intent when facing an error. Accordingly, asking a suitable question may guide a part of the users to find good rephrases to interact with the system.

%In the context of the example shown in Fig. 1, the predictor detects that the song \emph{show up} in the user utterance may be caused by an error from the system or user and gives a low confidence score to this candidate response. Accordingly, the system asks the user a clarification question \say{Okey, Do you wanna play the music \emph{show me love} from Jeff Kashiwa? And also, you could say it again}. This can circumvent user friction. Instead of correcting the user utterance directly, asking for clarification is a better way for the system to interact with the user when the system does not determine the real intent of the user. Moreover, some users in DuerOS tend to modify a previous utterance in hopes of fixing an error in the previous turn (which may be an error from the user or the system) to get the right results, but some users may give up or change the next utterance to another intent when facing an error. Accordingly, asking a suitable question guides a part of users to find good rephrases to interact with the system.

Based on the above reasons, we consider it is necessary to adopt such a proactive interaction mechanism in DuerOS, where one key challenge task is to decide whether to ask for clarification in each turn. A common solution of this task \cite{alok2020design, bodigutla2020joint, sekulic2021user} is to predict the user satisfaction with the candidate response that decides whether to ask for clarification in each turn. Though these works have been demonstrated their effectiveness in the offline datasets, it is hard to directly deploy them to DuerOS since more complex user dialogues that cover millions of domains make these methods hard to predict an accuracy user satisfaction in a commercial dialogue system. Furthermore, previous works usually adopt the turn-level user satisfaction to evaluate the user experience in a commercial dialogue system which is based on the information within the same turn. However, when deploying the proactive interaction mechanism to the DuerOS, turn-level user satisfaction cannot be adopted to evaluate the user satisfaction in our scenario since the evaluation should be based on the user utterance and the system response in the sequential turn as well. For example, we need to use the information of the sequential turn to determine whether the additional clarification question disturbs the user.   
 
The challenges we face in this paper are summarized as follows:
% In contrast to the previous work, there are three main challenges for our setting:

\begin{itemize}[leftmargin=*]
    \item \textbf{Labels are not sufficient.}
    In DuerOS, there exists only a few users or expert annotators to support a few labels on user sessions. Accordingly, how we can improve the model over time with a few labels is a challenge to us. 
    \item \textbf{Complex patterns lie behind different types of data.}
    % Extracting information Incorporating the information from structured and text data}
    The data available for our task is various, including structured data (such as the result item) and text data (such as the queries parsed by the ASR module). Moreover, the task handled by a commercial dialogue system is diverse and the data patterns behind these tasks become complex. 
    Accordingly, it is challenging to effectively extract the patterns from the data to predict the user satisfaction.
    \item \textbf{How to measure the user experience under the proactive interaction mechanism in DuerOS.}
     In DuerOS, we adopt the proactive interaction mechanism to improve the user experience. However, the evaluation of the user experience under this mechanism is complicated. When evaluating the user satisfaction in the turn when the system asks for clarification, we need to consider the user satisfaction with the system result in the sequential turn. 
\end{itemize}
To overcome these challenges, we propose a weak label generation method to generate weak labels. With the weak labels, we train a transformer-based model to predict the user satisfaction with the candidate response in the proactive interaction mechanism. In addition, we propose a new metric to evaluate the user experience under this mechanism. Specifically, the contributions in the paper are summarized as follows.

%we propose to generate a large number of weak labels automatically. 
%Then, inspired by the recent success of transformer-based models \cite{attention:2017}, 
%we present a new transformer-based model (see Figure \ref{fig:Model}) to discover the complex patterns and predict the confidence score that decides to whether ask for clarification on DuerOS. 
%Furthermore, we also propose to train the model using a large batch size which is faster and performs better when labels are noisy. To evaluate the effectiveness of our method, we conduct comprehensive experiments on three large datasets. The experimental results show that our 
%method achieves consistent improvements over the baseline methods. Finally, with a new designed evaluation method, we deploy and evaluate our model on DuerOS, which also brings significant improvement in user experience over the baseline methods. In summary, the contributions of our work are summarized as follows:
\begin{itemize}[leftmargin=*]
    \item First, we use a linear model with handcrafted features that are extracted from the user's interactions with the system in the current turn and the next turn to generate a large number of weak labels.
    \item Second, we propose a transformer-based model to extract the information from both the structured and text data and adopt a large batch size in the training to alleviate the problem induced by the noisy and conflicting labels.
    \item Then, we propose a new metric, contextual user satisfaction, to measure the user experience under the proactive interaction mechanism. 
    \item Finally, we show the effectiveness of our approach with a series of offline experiments on three large datasets and an online evaluation by deploying the model to DuerOS.
    % and deploy it to 
    % a real large-scale conversational system DuerOS.
\end{itemize}
 \begin{figure*}[h]
 	\centering\includegraphics[width=5.5in]{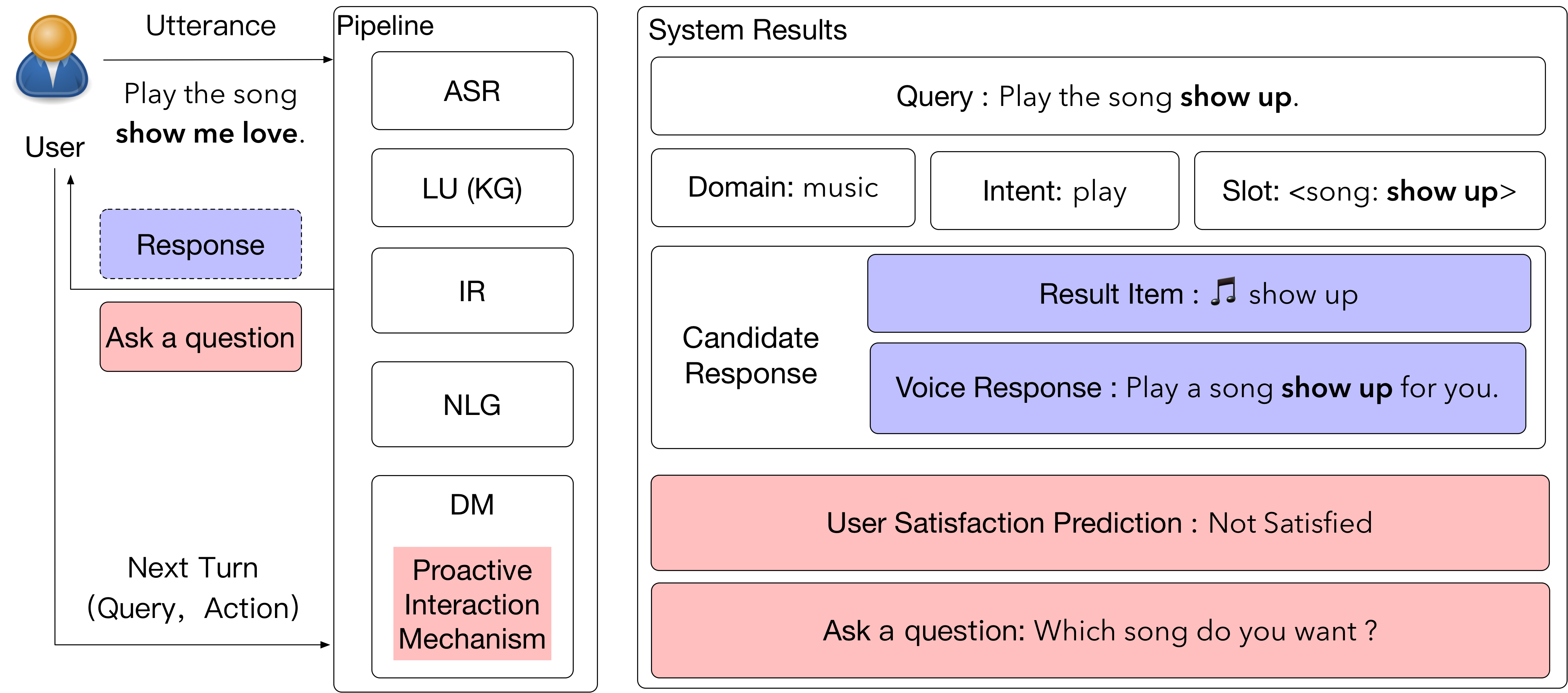}
 	\caption{
 The flow chart for the dialogue system.
 The system first generates a candidate response that consists of a result item and a voice response (the blue boxes).
 If the predictor predicts that the user is not likely to be satisfied with the candidate response, the system will ask a suitable question to determine the real intent of the user (the red boxes) instead of providing the candidate response to the user (the blue dotted box).
 	}\label{fig:Motivation}
\end{figure*}
\section{Related Work}
Our work touches on three strands of research: proactive interaction mechanism in spoken dialogue system, weak label generation process, and transformer-based model.
\subsection{Proactive interaction mechanism in spoken dialogue system}
 Proactive interaction mechanism is a widely studied research topic in spoken dialogue systems, which contains two key challenge tasks: \emph{When} (i.e., whether to ask for clarification in each turn) and \emph{how} (i.e., clarification query generation) to ask for clarification. There are many works focusing on the clarification query generation problem in the spoken dialogue systems \cite{severcan2020clarification, zamani2020mimics, majumder2021ask}. However, few researchers focus on whether to ask for clarification in each turn of the spoken dialogue systems, particularly in a commercial dialogue system. For example, \citet{hakkani2005error} propose a rule-based method to combine the signals from the ASR and LU module to predict whether to ask for clarification. In addition, \citet{kotti2017will} extract the features from ASR and LU module and train a linear model with provided labels to predict the user satisfaction with the candidate response to help the system decide whether to ask for clarification in each turn. Recently, \citet{alok2020design} propose a hypothesis rejection module that adopts a deep model with the user utterance and a series of handcrafted features from the NLU module to predict whether the system rejects the NLU result or directly give a response to the user in Alexa. When rejecting the NLU result, Alexa may ask for clarification or give no response to the user. However, the models designed by the previous works only focus on the information in the current turn and ignore the rich information in the user sessions, which may reveal the user preference and potential intention (e.g., whether a user prefers a short video or not). Moreover, most of the previous works are evaluated on the offline datasets, and not deployed to a real dialogue system. 
 In our work, we design a transformer-based model to exploit the complex patterns from both the current utterance and the user sessions to predict whether to ask a clarification question in the spoken dialogue system. Moreover, we deploy our model to a commercial dialogue system, DuerOS, and evaluate not only the performance of the model but also the user experience under the proactive interaction mechanism by a new metric.

\begin{figure*}[!t]
 	\centering\includegraphics[width=5.5in]{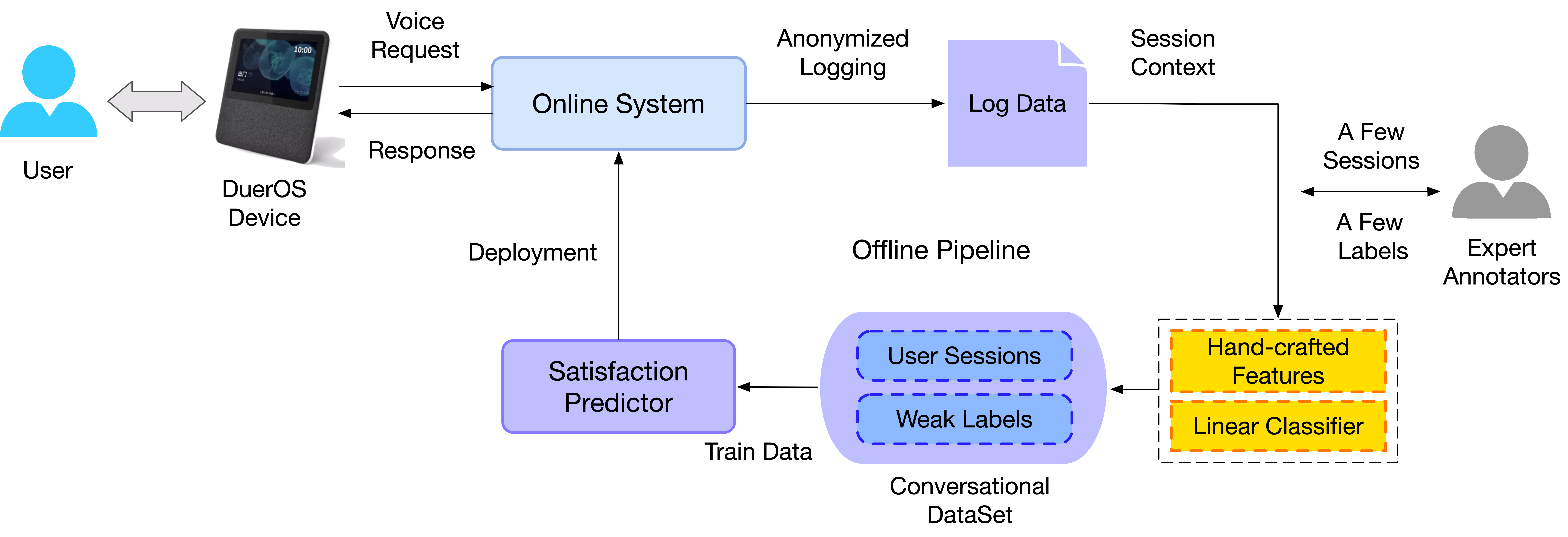}
 	\caption{A high-level overview of our method. There are two stages in our solution to the user satisfaction prediction problem: 
 	 1) Weak label generation process: We first collect anonymous user log data 
%  	 from a daily basis 
    and turn-level user satisfaction labels annotated by humans.
    % with them from expert annotators. 
    Based on these data and handcrafted features, we train a linear model to generate weak labels for unlabeled data. 
 	 2) User satisfaction prediction process: We use dialogue-level features from user sessions and generated weak labels to train a transformer-based model to predict user satisfaction with the candidate response. Finally, we deploy the user satisfaction predictor to the online system that decides whether to ask a question for clarification in each turn.
 	}\label{fig:overview}
 \end{figure*}
 
\subsection{Weak Label Generation Process}
To automatically generate the weak labels of the user satisfaction with the candidate response, we estimate the user satisfaction with the system response in each turn. Different from the user satisfaction prediction with the candidate response, when estimating the user satisfaction with the system responses, we can use the user utterance and user actions after the current turn, which contains the information of the user feedback to the system response in the current turn.

In the previous work, user satisfaction is a subjective measure of a user's experience in a dialogue system, which indicates whether the user’s desire or goal is fulfilled. There is two-level user satisfaction in spoken dialogue systems: dialogue level user satisfaction which is to estimate whether the system response satisfies the user at the end of a user dialogue or session, and turn level user satisfaction which is to estimate whether the system response satisfies the user in each turn. To evaluate the turn-level user satisfaction on a spoken dialogue system or intelligent assistants, \citet{hashemi2018measuring} first propose to extract the user intent from the user utterances and embed it into the query embeddings. Based on the intent-sensitive query embeddings, they measure the turn-level user satisfaction. In addition, \citet{bodigutla2019multi} proposes a new response quality annotation scheme, which introduces several domain-independent features to estimate user satisfaction, which improves generalizability to conversations spanning over multiple domains. Furthermore, \citet{bodigutla2020joint} propose a BiLSTM based model that automatically weighs each turn’s contribution towards the dialogue-level user satisfaction to jointly estimate the turn-level and dialogue-level user satisfaction. 

In our work, we adopt a linear model with cross-domain hand-crafted features to predict turn-level user satisfaction, which is used as the weak label of our model. In other words, we assume that if a user is not satisfied with the system response in one turn, proactively interacting with the user may bring a better experience to the user than directly giving this response. Moreover, a large number of weak labels enable us to train a deep neural network to explore more complex patterns from the user sessions that predict the user satisfaction with the candidate response.

\subsection{Transformer-based model}
Transformer is first proposed by \citet{attention:2017}, which is solely based on an attention mechanism.
It is more parallelizable and therefore being more computationally efficient compared with RNN or CNN.
Transformer is effective for a wide range of problems.
The most notable success is BERT \cite{bert:2018} which is a transformer-based pre-trained model that works well on a wide range of natural language processing (NLP) tasks after fine-tuning, such as question-answering and natural language inference.
Transformer-based models are also effective for relational reasoning \cite{zambaldi2018deep}, multi-agent reinforcement learning \cite{vinyals2019alphastar}, and improving the click-through rate (CTR) in recommendation systems \cite{chen2019behavior}.
Notice that our task is closely connected to NLP (since there is text data in the dialogue system) and CTR prediction (since we need to predict the preference of the user). Accordingly, we design a transformer-based model to handle text data and structure data separately and adopt a self-attention module to extract temporal dependency between the current turn and the previous turns. With the complex patterns and temporal dependency learned in the user sessions, our transformer-based model outputs the user satisfaction with the candidate response to help the system decide whether to ask for clarification in each turn. 
 
\section{System Overview}
The DuerOS understands a user utterance by a pipeline of the following modules: 
automatic speech recognition (ASR) module, language understanding (LU) module, 
knowledge graph (KG) module, 
information retrieval (IR) module, natural language generation (NLG) module,
dialogue manager (DM) module and text-to-speech (TTS) module. In DuerOS, we deploy the proactive interaction mechanism to the DM module.

Given an utterance from a user (i.e., a voice command), 
the system goes through the following steps to generate a response (see Figure \ref{fig:Motivation} Left):
% 1) the ASR module converts the utterance into a text;
1) the ASR module converts the utterance into a query (i.e., a text);
2) the LU module recognizes the domain and intent for the given query, and fills the associated slots for the domain-specific semantic template; 
% 4) the IR module and NLG module generate a candidate response, a suitable item, and voice response, according to the domain-intent and slots.
3) the IR module and NLG module generate a candidate response composed of a suitable result item and a voice response, according to the domain-intent and the slots. Generally, a voice response is a piece of audio from the TTS module. For simplicity, we use voice response to refer the input text of the TTS module;
4) the DM module gives the candidate response associated with the prediction of user satisfaction. Then, according to the user satisfaction, the DM module decides whether to give a candidate response or ask a clarification question to the end user.

For example (see Figure \ref{fig:Motivation} Right), 
% Let us look at an example shown in Figure \ref{fig:Motivation}: 
a user gives an utterance
``Play the song \emph{show me love}'' to the system. 
If the ASR module makes a mistake and converts the voice into a wrong text ``Play the song \emph{show up}'',
the LU module may parse the text into \texttt{(domain-intent: music-play, slots: $\langle$song: \emph{show up}$\rangle$)}, which is incorrect.
Then, the IR module and the NLG module generate a wrong candidate response correspondingly. In the DM module, we deploy a predictor to predict the user satisfaction with this response (which is between $0$ and $1$). When the predicted user satisfaction is smaller than the threshold $T$, the DM module would decide to ask a clarification question to the end user instead of providing the wrong candidate response.   
 \begin{table*}[t]
    \caption{Features used for generating the weak labels}
	\centering
	\small
% 	\begin{tabular}{m{50pt}<{\centering} m{50pt}<{\centering} m{40pt}<{\centering} m{40pt}< {\centering}}
	\begin{tabular}{cll}
    	\hline
		Index & Feature set description & Turn(s)\\
		\hline
1 & ASR Confidence & $I_n$ \\
2 & Avg time difference between consecutive utterances & $I_n$, $I_{n+1}$ \\
3 & Affirmation prompt in user utterance & $I_n$ \\
4 & Affirmation prompt in next turn’s user utterance & $t_{n+1}$ \\
5 & Negation prompt in user utterance & $I_n$ \\
6 & Negation prompt in next turn’s user utterance & $I_{n+1}$ \\
7 & Domain popularity computed on predicted NLU intent & $I_n$ \\
8 & Domain popularity computed on next turn’s predicted NLU intent & $I_{n+1}$ \\
9 & Intent popularity computed on predicted NLU intent & $I_n$ \\
10 & Intent popularity computed on next turn’s predicted NLU intent & $I_{n+1}$ \\
11 & Length of the utterance & $I_n$, $I_{n+1}$ \\
12 & Next turn’s ASR Confidence & $I_{n+1}$ \\
13 & Next turn’s NLU Confidence & $I_{n+1}$ \\
14 & NLU Confidence & $I_n$ \\
15 & NLU Intent similarity between consecutive turns & $I_n$, $I_{n+1}$ \\
16 & Syntactic similarity between consecutive turns user utterances & $I_n$ \\
17 & Syntactic similarity between current response and next turn’s system response & $I_n$, $I_{n+1}$ \\
18 & Syntactic similarity between current response and previous turn’s system response & $I_n$ \\
19 & Syntactic similarity between user utterance and system response & $I_n$ \\
20 & Termination prompt in user request & $I_n$ \\
21 & Termination prompt in next turn’s user request & $I_{n+1}$ \\
		\hline
	\end{tabular}
	\label{table:Features}
\end{table*}
\section{Method}
We show a high-level overview of our method in Figure. \ref{fig:overview}. Specifically, there are two steps in our solution to predict the user satisfaction with the candidate response: 1) Weak label generation process: We first collect a small amount of anonymized user log data with the labels annotated by the expert annotators \footnote{Expert annotators consistently achieve a high agreement (correlation is bigger than 0.8) with other expert annotators and with explicit turn-level user ratings collected through DuerOS app.}. Based on these data and handcrafted features, we train a linear model to generate the weak labels.   2) User satisfaction prediction process: We use dialogue-level features in user sessions and a large number of weak labels to train a transformer-based model to predict the user satisfaction with the candidate response.
Finally, we deploy the user satisfaction predictor to the DM module in DuerOS to decide whether to ask for clarification or give the candidate response to the user in each turn.    
\subsection{Problem definition}
Here, we formally define the problem and introduce the notations.
The user satisfaction prediction problem in the spoken dialogue system is to use the dialogue information to predict the user satisfaction with the candidate response that helps the system decide whether to ask for clarification or give the candidate response to the user in each turn. 

Similar to \citet{zhang2019joint}, we define a dialogue turn at time $n$ as $t_n = (q_n, d_n, s_n, r_n, v_n)$. The query $q_n$ is the decoded text of the user utterance; $d_n$ is the domain-intent; $s_n$ is the slots; $r_n$ is the system results; and the voice response $v_n$ is a text serving as the input to a subsequent text-to-speech (TTS) module.
Moreover, we denote the corresponding user satisfaction with the candidate response ($r_n$, $v_n$) as $l_n$. Our goal is to train a predictor that predicts $l_n$ based on $(t_1,t_2,...,t_n)$.
In addition, we use the interactions between the user and the system in the current turn and the next turn $(I_{n}, I_{n+1})$ to infer the user satisfaction \emph{ex post} through a weak label generation process.
% The probability that the user is satisfied $\hat{l}(a,q')$ can be processed to be a weak label $\hat{l}$.

% Our solution to the user satisfaction prediction problem is composed of a weak label generator and a user satisfaction predictor.
% In the stage of weak label generation, we adopt handcrafted features extraction, a logistic regression model, and noise reduction to generate the weak labels. 
% In the stage of user satisfaction prediction, we train a transformer-based model with large-batch training.  

\subsection{Weak label generation}

In this stage, we generate weak labels for all the turns based on a few labeled turns from the expert annotators via a logistic regression model. The features of the current turns listed in the Table \ref{table:Features} are extracted from the current turn and next turn (i.e., $I_n$ and $I_{n+1}$). 
The examples of the features are \emph{Avg time difference between consecutive utterances} (which suggests that the user gets a satisfying response if the time is long) and \emph{Negation prompt in next turn’s user utterance} (e.g., certain words in the next turn's utterance indicate that the user is unhappy).

Based on the features and a few labeled samples, we train a logistic regression model.
Specifically, we learn the posterior probability of the event that the user is satisfied $\hat{l_n} := p(l_n|I_{n},I_{n+1})$ and use it as the weak label.
Notice that this is an inference for user satisfaction \emph{ex post}, which is different from and certainly simpler than the predictor we train in the next stage.
To this end, a logistic regression model is sufficient to fit the data.

\subsection{User satisfaction prediction with transformer-based model}
\begin{figure*}[!t]
    \centering\includegraphics[width=6.5in]{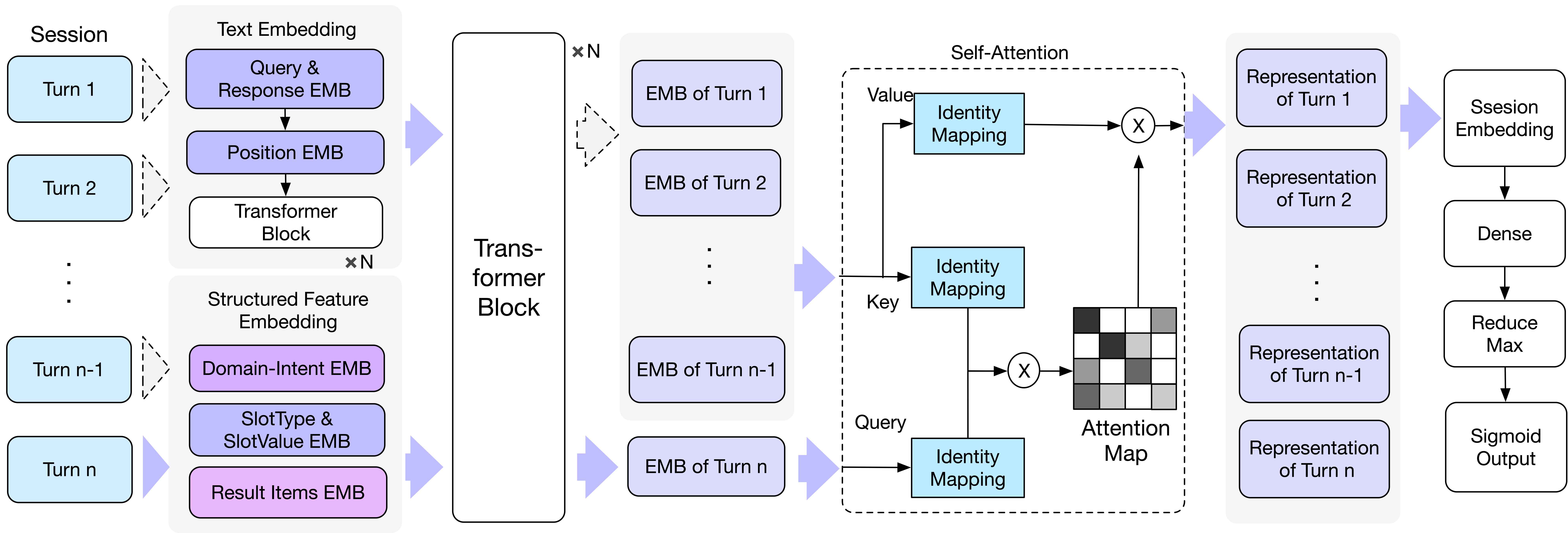}
    \caption{
    % The architecture of the transformer-based model.
    The architecture of the transformer-based model.
    }
    \label{fig:Model}
\end{figure*}

With a large number of samples and the weak labels generated in the previous stage, we train a transformer-based predictor to predict the user satisfaction with the candidate response $l_n$ based on the features $(t_1, t_2, ..., t_n)$.
The architecture of the transformer-based model is shown in Figure \ref{fig:Model}.
The model contains two parts: (1) In the first part, the model adopts two stacks of transformer blocks to learn the patterns in each turn, where the first stack is used to extract features from the interactions between the query and the voice response which are text features, and the second stack is to extract features from the interactions between the domain-intent, the slot, the result item, and the query-response. The output of the first part is the embedding vector of each turn. (2) In the second part, the model adopts scaled dot-product attention \cite{attention:2017} to extract features from the interactions between the current turn and the previous turns.

In the first part, the query and the voice response are processed by a query-response embedding layer followed by a position embedding layer, which follows \citet{attention:2017}.
Then, the embeddings are fed to the first stack of transformer blocks, each of which consists of a multi-head attention layer, a layer-norm layer, a feed-forward layer, and another layer-norm layer (cf. Figure \ref{fig:Model} Center).
The first stack of transformers block models the interaction between
the query and the voice response.
% each item in the query and each item in the response.
This architecture is helpful for the predictor to identify the mismatch between the query and the voice response which may indicate an erroneous response and the dissatisfaction of the user. Then, we use the second stack of transformer blocks with a similar architecture to extract the error patterns from the interactions between the structured features (the slot $s_i$, the domain-intent $d_i$, the item $m_i$) and the output vectors of the first stack of transformer blocks  (i.e, the representation of the query $q_i$ and response $r_i$). Finally, the output of the first part is the embedding vector of each turn $E_i$.

In the second part, the input of the scaled dot-product attention consists of \emph{query}, \emph{key} and \emph{value}, where \emph{query} is the embedding vector of the current turn (denoted by $Q$), \emph{key} is the embedding vectors of the previous turns (denoted by $K$), and \emph{value} is also the embedding vectors of the previous turns (denoted by $V$). The output (denoted by $O$) of attention is a weighted sum of the value, where the weight matrix is determined by \emph{query} and its corresponding \emph{key}.
\begin{equation}
    O = softmax(\frac{QK^T}{\sqrt{d}})V
\end{equation}
where $d$ is usually set to a large value in our case to scale the dot product attention. 

At last, we use a fully-connected layer followed by \emph{reduce max} function to find the error patterns in the interactions between the current turn and the previous turns. With the \emph{sigmoid} function, the model outputs the probability $p(l_n|t_1,t_2, ...,t_n)$. In addition, we use the cross entropy between the prediction $p(l_n|t_1,t_2, ...,t_n)$ and the weak label $\hat{l}$ as the loss function.

\subsection{Large-batch training}
There are two potential problems in the above training process with weak labels: 
1) Training a transformer-based model with a large amount of data is time-consuming. 
Moreover, when deployed online, the model needs to be frequently retrained (e.g., for every two weeks) using the updated data, which places a higher requirement of training efficiency.
Retraining is necessary since the system is consistently updated in practice (e.g., the change of system settings or the update of the knowledge base).
The update may result in a shift of the data distribution and degrade the performance of the old predictor.
% In our system, the change of the system settings and the knowledge base results in the change of the system \say{errors}. For grasping these up-to-date \say{errors}, it is necessary to retrain the model with new samples in a period of time. This places a higher requirement of training efficiency.
2) The noise in the training data (induced from the weak label generation) slows down the learning and affects the generalization of the model. Specifically, for gradient-based methods, the noisy labels may lead to the wrong direction of the gradient.
The model optimized through successive erroneous gradient steps may lead to a poor local minimizer and therefore poor generalization. 

To solve the above problems, we propose to use a large batch size 
% large-batch training 
(LB) in the training process. For the first problem, LB accelerates the training process especially when the model is trained on multiple machines.
LB reduces the number of passes through the model to iterate over all the training samples once.
Also, LB is more suitable for parallelization across multiple machines. For the second problem, too many noisy samples in a batch may result in an incorrect gradient update, and successive erroneous gradient updates may lead to a deviation from the gradient updates using the \emph{true} gradient which is assumed to result in correct generalization. Here, we use \emph{true} gradient to refer to the gradient calculated on the samples with correct labels. In the training process,
if the batch size is large enough, the noise ratio (i.e., the percentage of samples with incorrect labels) in a batch will not be too large and thus the gradient will be close to the \emph{true} gradient with large probability. This may result in a better generalization.
In general, we find that LB leads to better generalization when there are noisy samples.

\section{Measure User Experience in DuerOS}
\label{Evaluation}
 Though DuerOS collects the feedback from users by an app, the amount of the feedback is too few to be adopted to evaluate the performance of the model and the user experience. Previous works usually adopt turn-level user satisfaction (TUS) to measure user experience in a commercial dialogue system. However, TUS cannot be adopted to evaluate the user satisfaction in our scenario since the evaluation should be based on the user utterance and the system response in the sequential turn as well. Accordingly, we propose a new metric, contextual user satisfaction (CUS), to evaluate the user experience in our scenario. Specifically, we first evaluate the TUS by two ways: 1) use the linear model to automatically rate each turn on a continuous $[0-1]$ scale, 2) rate some specific turns on a discrete $[0-1]$ scale from expert annotators. For the above example in Figure. \ref{fig:Motivation}, the linear model would rate the turn ``Hey Xiaodu, play the song \emph{show me love}'' as a score $0.87$, while the expert annotators rate this turn as $1$, which indicates that the user is satisfied with the system response and has a good experience. In contrast, if the system misrecognizes the user utterance as "play the song \emph{show up}" in this turn and plays the song \emph{show up}, the linear model rates this turn as a score $0.4$ and the expert annotators rate this turn as $0$, which indicates the system does not fulfill the user requirement. 
 Then, we use TUS to calculate CUS by taking account of the contextual information around the turn when the system ask for clarification. Specifically, when the system proactively asks a question, there are two steps to estimate the user experience. First, we estimate whether the user is satisfied with the clarification question. Second, we estimate whether the new response from the system satisfies the user. When the user is not satisfied with the clarification question or the system response after the clarification, we consider the question disturbs the user. Accordingly, we estimate the user experience (denoted by $r_c$) when the system asks for clarification by the following equation,
 \begin{equation}
     r_c = r_n * r_{n+1}.
 \end{equation}
 where $r_n$ is the rating of the user satisfaction in the turn when the system asks a clarification and $r_{n+1}$ is the rating of user satisfaction in the turn after the clarification.
% As shown in Figure \ref{fig:Gradient}, the trajectory of the parameters trained by LB has a large probability to be \say{close} to the trajectory of the parameters trained by the \say{true} gradient. Therefore, the model parameters trained by LB is not too far away from the model parameters trained by the \say{true} gradient.
% However, the trajectory of the parameters trained by SB may be too far away from that trained by the \say{true} gradient, which results from the completely wrong gradient in some steps. It may result in a poor generalization. Generally speaking, LB ensures the optimization and generalization of the model. In Appendix A.2, we give a simple theory on the relationship between the model parameters trained by LB and the \say{true} gradient and its proof.

%However, large batch size training has some drawbacks, 
%\begin{itemize}
%	\item Generalization Problem, large batch size training loss its accuracy in test set though it is able to obtain high accuracy in training set \cite{}.
%	\item Optimization Difficulty, it is hard to get right hyper-parameter to optimize the model \cite{}.
%\end{itemize}

%Some techniques have been adopted in previous work to solve these problems, such as warming up learning rate \cite{}, layer-wise adaptive rate scaling \cite{} and Layer-wise Adaptive Moments for Batch (LAMB) \cite{}.
%Due to the transformer-based model, we adopt LAMB optimizer to optimize the model in this paper.
\begin{table*}[t]
	\centering
	\caption{Scales of the three datasets and the statistics of the weak labels.}
% 	\begin{tabular}{m{120pt}<{\centering} m{80pt}<{\centering} m{80pt}<{\centering} m{80pt}<{\centering}}
    \begin{tabular}{l|rrr}
		\hline
		 & Dataset 1 &  Dataset 2 & Dataset 3\\
		\hline
		\#samples in training set & 18M &  32M & 5M \\ 
		\#samples in validation set & 250K & 250K & 250K\\
		\#samples in testing set & 1K  & 1K  & 1K \\
		Accuracy of the weak labels  & 0.833 & 0.832 & 0.846\\
		Performance of our model on the validation sets (AUC) & 0.956 & 0.942 & 0.963\\
		Performance of our model on the testing sets (AUC) & 0.810 & 0.797 & 0.754 \\
		\hline
	\end{tabular}
	\label{table:dataset}
\end{table*} 

\begin{table*}[t]
	\centering
	\caption{The performance comparison between the baselines and our model on testing sets.}
% 	\begin{tabular}{m{120pt}<{\raggedright} m{80pt}<{\centering} m{80pt}<{\centering} m{80pt}<{\centering}}
	\begin{tabular}{r|ccc}
		\hline
		Model & Dataset 1 (AUC / CLA) &  Dataset 2 (AUC / CLA) & Dataset 3 (AUC / CLA)\\
		\hline
		Trans-Text & 0.724 / 0.050& 0.668 / 0.045& 0.643 / 0.020\\
		LSTM & 0.765 / 0.060 &  0.742 / 0.058 & 0.737 / 0.041\\ 
		CNN &  0.734 / 0.051 & 0.742 / 0.042 & 0.725 / 0.048\\
		TDU & 0.754 / 0.071 & 0.787 / 0.052 & \textbf{0.767} / 0.045\\
		TBM (ours)  & \textbf{0.810} / \textbf{0.085} & \textbf{0.797} / \textbf{0.060} & 0.754 / \textbf{0.053}\\
		\hline
	\end{tabular}
	\label{table:performance_compare}
\end{table*} 
\iffalse
\begin{table}[t]
%	\renewcommand{\arraystretch}{1.3}
	\centering
	\caption{The online testing results.}
	\begin{tabular}{m{80pt}<{\centering} m{80pt}<{\centering}}
		\toprule
		&  Online Result \\
		\midrule
		Precision & 0.900\\ 
		Recall & 0.160\\
		Conversion Ratio & 0.675\\
		\hline
	\end{tabular}
	\label{table:online_result}
\end{table}
\fi

\iffalse
\begin{table}[t]
%	\renewcommand{\arraystretch}{1.3}
	\centering
	\caption{We evaluate how noisy reduction, time and weak labels affect the performance on the testing data.}
	\begin{tabular}{m{30pt}<{\centering} m{40pt}<{\centering} m{40pt}< {\centering}}
		\toprule
		& AUC & Accuracy\\
		\midrule
		Train & 0.970 & 0.968\\
		Validate & 0.960 & 0.956\\ 
		NR\_Val & 0.956 & 0.945\\
		Time\_Val  & 0.958 & 0.948\\
		Test & 0.810 & 0.720\\
		\hline
	\end{tabular}
	\label{table:experiment1}
\end{table}
\fi

%加入图信息，解释时间，增加时长
\section{Experiment}
In this section, we conduct experiments on three real datasets and evaluate our model online on DuerOS. 
We aim to answer the following questions:

\textbf{Q1:} How does the weak label generator perform?

\textbf{Q2:} How does our transformer-based model perform compared with the other baselines on the three datasets?

\textbf{Q3:} How do different designs influence the performance of our model?

\textbf{Q4:} How does training using a large batch size (LB) perform compared with training using a small batch size (SB)?

\textbf{Q5:} How does our transformer-based model perform when we deploy it on DuerOS?

%\textbf{Q4:} Which factor (‘time’, noisy reduction module, and weak labels) affects the test results based on the feedback from the real users most?

%\textbf{Q5:} Does the model has a good generalization ability and memorization ability? 
\subsection{Offline experiments}
\subsubsection{Dataset}

We conduct experiments on three industrial datasets of different scales, which are collected from DuerOS.
% We conduct experiments on three industrial datasets of different scales, which are industrial datasets derived from a large-scale dialogue system (DuerOS).
The training and the validation data of the three datasets are extracted from 
% the user data in
different months (June, July, and August respectively). 
% The scale of the three datasets is shown in Table \ref{table:dataset}.
The training data of the three datasets contains 18M, 32M, and 5M samples respectively.
The validation data of each dataset contains 250K samples. 
The training data and the validation data are all labeled by the weak label generator. The testing data contains 1k samples and are labeled by expert annotators, which can be regarded as the ground truth. 
For clarity, we summarize these statistics in Table \ref{table:dataset}.

% \begin{table*}[tp]
% 	\centering
% %	\fontsize{11}{8}\selectfont
%     \caption{Large-batch training (LB) and small-batch training (SB) performance comparison. }
% % 	\begin{threeparttable}
% 	\begin{tabular}{ccccccc}
% 		\toprule
% 		\multirow{2}{*}{Epoch}&
% 		\multicolumn{3}{c}{SB (batch\_size = 1024)}&\multicolumn{3}{c}{LB (batch\_size = 12000)}\cr
% 		\cmidrule(lr){2-4} \cmidrule(lr){5-7}
% 		&Loss&Train AUC&Testing AUC&Loss&Train AUC&Testing AUC\cr
% 		\midrule
% 		0&3.3&0.50& 0.50 &3.3&0.50& 0.50\cr
% 		20&2.6&0.51& 0.50 &\textbf{1.8}&0.97& \textbf{0.81}\cr
% 		40&2.5&0.63& 0.54 &1.8&0.98& 0.80 \cr
% 		60&2.0&0.72& 0.63 &1.8&0.98& 0.77 \cr
% 		80&\textbf{1.8}&0.97&\textbf{0.79}&1.8&0.99& 0.78\cr
% 		100&1.8&0.99&0.77&1.8&0.98& 0.79\cr
% 		\bottomrule
% 	\end{tabular}
% % 	\end{threeparttable}
%     \label{table:performance_comparison}
% \end{table*}

\subsubsection{Metrics and hyperparameters}
In our experiment, we use the area under the curve (AUC) and the conditional label accuracy (CLA) as the evaluation metrics. 
AUC is the area under the ROC curve, and 
CLA is the maximum recall when precision is larger than a specified value (which is set to 85\% in this paper).
Accuracy is the percentage of correct predictions. In our experiments, we binary our prediction according to a threshold and then calculate the classification accuracy given the true labels, where we tune the threshold by grid searching to maximize the accuracy.

We train the weak label generator with $10$k training samples from the expert annotators. Then we implement our model based on Tensorflow and use the Adam optimizer. We apply a grid search for hyperparameters: the number of layers in two transformer blocks is searched in $\{1, 2, ..., 8\}$ and the embedding size of each vector in the embedding layer is chosen from $\{90, 120,..., 1800\}$. Besides, the number of turns in our model is chosen from $\{1, 2,..., 10\}$. (cf. Figure \ref{fig:Model}). The selected hyperparameters of the model are listed in the Tabel \ref{table:Hyp}. 

\begin{table}[!h]
    \caption{Hyperparameters in the model}
	\centering
	%\begin{tabular}{p{200pt}|m{80pt}}
	%\begin{tabular}{p{8cm} p{2.5cm}<{\centering}
	\begin{tabular}{lr}
	%\begin{tabular}{|p{6cm}|m{0.5cm}<{\centering}|}
       	\hline
	    Hyperparameters & Value\\
		\hline
 LR & $1.2\times 10^{-3}$ \\
 The number of turns & $5$ \\
 Embedding size & 240\\
 Threshold & 0.7\\
 The numbers of the first stack of the transformers & $8$ \\
 The numbers of the second stack of the transformers & $4$ \\
		\hline
	\end{tabular}
	\label{table:Hyp}
\end{table}

\subsubsection{Baselines}
% In order to verify the validity of Transformer-based model, we compare it with the following baselines:
We compare our transformer-based model (TBM) with the following baselines:
\begin{itemize}[leftmargin=*]
    \item \textbf{Trans-Text.} Trans-Text \cite{hakkani2005error} is a simple but efficient user satisfaction prediction method based on prescribed rules. This method takes the semantic similarity and time interval between two turns, word confidence in the query, and frequency of occurrence of the user queries into consideration.
    \item \textbf{TDU.} TDU \cite{zhang2019joint} is a BiLSTM based deep neural net model in Alexa that predicts both turn-level and dialogue-level user satisfaction.
    \item \textbf{CNN and LSTM.} 
    We also implement several popular architectures: CNN (following Text-CNN in \citet{kim2014convolutional}) and LSTM \cite{hochreiter1997long}.
    We train the models with these architectures using the same training sets (including the weak labels).
\end{itemize}

\subsubsection{Results}

In the subsequent paragraphs, we first present the performance of the weak label generator. Then, we present the performance comparison between the baselines and the transformer-based model on the three datasets, and an ablation study of our model. 
At last, we compare the performance of LB and SB on Dataset 1.
% At last, we conduct an A/B test on a large-scale spoken dialogue system (DuerOS) to test the effectiveness of the transformer-based model.

\textbf{The performance of weak label generator (Q1).}
We show the accuracy of the weak labels from the generator on the three datasets in Table \ref{table:dataset}. 
We have the following observations on the experimental results:
\begin{itemize}[leftmargin=*]
    \item Based on the available ground truth, we can evaluate the accuracy of our weak label generator, which is around 84\% for the three datasets. It demonstrates that the weak label generator performs well on the three large datasets.
    \item 
    When trained using the generated weak labels,
    the transformer-based model obtains good performance in the three validation sets (with weak labels).
    This demonstrates the weak labels generalize well across the data. 
    % When trained by a large amount of the weak labels, transformer-based model can obtain good performance on three valid datasets. It demonstrates that our model has a good generalization ability. 
    \item 
    However, we observe that the performance of our model on the testing sets (with true labels) is slightly worse than that on the validation sets. 
    This demonstrates the gap between weak labels and true labels, which motivates us to further improve the accuracy of the weak labels in the future.
    % really have a bad effect on our model and we need to improve the accuracy of the weak labels in the future. 
    % However, the performance of the transformer-based model on test datasets is worse than on valid datasets. It demonstrated that the weak labels really have a bad effect on our model and we need to improve the accuracy of the weak labels in the future. 
\end{itemize}
 
\textbf{Performance comparison (Q2).}
We show the experimental results on the three datasets in Table \ref{table:performance_compare}. 
We have the following observations on the experimental results:
% We have the following observations about the results of the error prediction task:  

\begin{itemize}[leftmargin=*]
    \item Transformer-based model outperforms all the other baselines on Dataset 1 and Dataset 2. 
    In particular, compared with TDU, the performance of our model has improved by 7.4\% in AUC (or 19\% in CLA) on Dataset 1, and 1.2\% in AUC (or 15\% in CLA) on Dataset 2.
    % improves the strongest baseline Bidirectional LSTM by 7.4\% and 19\% in AUC and CLA on dataset 1, and 1.2\% and 15\% in AUC and CLA on dataset 2. 
    The success of our model is due to the design of the transformer-based model that effectively extracts the information from structured and text data by modeling the interactions between the features and learns the temporal dependencies between turns by modeling the interactions between the features in current turn and those in previous turns.
    % We attribute such notable improvements to the powerful design of transformer-based model that can benefit from both explicit and implicit interactions and representation fusions among both text and structured features.   
    \item On Dataset 3, the AUC of the transformer-based model is slightly worse than that of TDU. 
    However, the CLA of our model is still better than other baselines.
    Notice that the number of samples in Dataset 3 is smaller than that of Dataset 1 and Dataset 2, which indicates that our model is suitable for scenarios with a large number of training samples.
    \item The rules in Trans-Text are carefully designed by experts
    % hand-crafted 
    and the model is not trained using data (and therefore weak labels).
    This model is shown to be effective in practice and being deployed on many real dialogue systems (e.g., AT\&T Spoken Dialog System).
    Nevertheless, Trans-Text is not as good as the other baselines.
    % Unexpectedly, Trans-Text is worse than all other baselines. Since the Trans-Text is not trained by a large amount of weak labels and each feature in this model is hand-crafted,  it is always effective when deployed on many real dialogue systems (e.g. AT\&T Spoken Dialog System). 
    % As such,
    % it demonstrates that the feature engineering of our model and weak labels generator is effective on large-scale datasets. 
\end{itemize}

\begin{table*}[t]
    \caption{Ablation study.}
	\centering
% 	\begin{tabular}{m{50pt}<{\centering} m{50pt}<{\centering} m{40pt}<{\centering} m{40pt}< {\centering}}
	\begin{tabular}{ccc|cc}
    	\hline
		\#Turns & Structured Transformer & Text Transfromer & AUC & Accuracy\\
		\hline
		T = 5 & N = 4 & N = 8& 0.810 & 0.720\\
		\hline
		T = 5 & Null & N = 8& 0.698 & 0.710\\ 
		T = 5 & N = 4 & Null & 0.678 & 0.700\\
		T = 5 & N = 4 & N = 4 & 0.773 & 0.710\\
		T = 5 & N = 8 & N = 8 & 0.811 & 0.718\\
		\hline
		T = 4 & N = 4 & N = 8 & 0.780 & 0.704\\
		T = 5 & N = 4 & N = 8 & 0.810 & 0.724\\
		T = 10 & N = 4 & N = 8 & 0.809 & 0.722\\
		T = 20 & N = 4 & N = 8 & 0.808 & 0.721\\
		\hline
	\end{tabular}
	\label{table:ablation study}
\end{table*}

\iffalse
 \begin{table}[t]
	\caption{Hyperparameters and training time of the model using a large batch size (LB) and a small batch size (SB).
% 	large-batch training (LB) and small-batch training (SB)
    }
    \label{table:LB_params}
	\centering
% 	\begin{tabular}[lrr]
% 	\begin{tabular}{m{110pt}<{\raggedright} m{40pt}<{\centering} m{40pt}< {\centering}}
	\begin{tabular}{l|cc}
	    \hline
		Hyperparameters & LB & SB\\
		\hline
		Learning rate & 0.012 & 0.001\\
		Batch size & 12,000 & 1,024\\ 
% 		optimizer & Adam & Adam\\
		Wall time for one batch  & 0.22s & 0.13s\\
		Total training time & 1.83h & 50.8h\\
		\hline
	\end{tabular}
\end{table}
\fi

\textbf{Ablation study (Q3).}
Here, we study the role of two stacks of transformer blocks, how the number of the transformer blocks in the two stacks (i.e., the number of the transformer blocks for \emph{text} data and the transformer blocks for \emph{structured} data) in our model affects the performance. Moreover, we also explore the temporal dependency between turns in the model that affects the performance. 
We perform the ablation study on Dataset 1 and show the results in Table \ref{table:ablation study}.
We have the following observations:

\begin{itemize}[leftmargin=*]
    \item First, we train the model with only the first stack of transformer blocks (for text data) or the second stack of transformer blocks (for structured data) separately. 
    The AUC and the accuracy of the two ablated versions with a single stack are lower than the transformer-based model. 
    This demonstrates that extracting the information from both the structured and text data using the transformer-based model improves the performance.
    \item We also test the model with the different number of transformer blocks (using $N=4$ or $8$).
    We find that a shallow structure for the first stack of transformer blocks (for text data) degrades the performance while a shallow structure for the second stack of transformer blocks (for structured data) does not significantly influence the performance.
    \item With the different number of turns in a user session fedding to the model, We find that when the number of the user turns is smaller than $5$ in TBM, the performance of the model degrades. However, when the number of the user turns is more than $5$ in TBM, there is no significant performance improvement of the model. It indicates that the user satisfaction in the current run has a strong relationship with the last four user turns. In other words, the temporal dependency between the current turn and the last four turns may exist and can be used to predict the user satisfaction with the candidate response. It may be caused by the reason that the user always has a similar intention in a continuous dialogue. 
    % This indicates that 
    % \item In addition, we test the model with the different number of transformer blocks in the first and the second transformer layer.
    % We find that when we reduce the number of transformer blocks in first transformer layer to 4, the performance declines. When we increase the number of transformer blocks in second transformer layer to 8, the performance has no notable improvement. As such, it is necessary to stack at least eight transformer blocks in the first transformer layer and stacking too much (larger than 4) layers in second transformer layer will not bring a big promotion.  
\end{itemize}

\textbf{The performance comparison between LB and SB (Q4).}
We compare LB and SB on Dataset 1.
For LB, we set the batch size to $12000$ and the learning rate to $0.012$.
For SB, we set the batch size to $1024$, and the learning rate to $0.001$.
The optimizers for both SB and LB are Adam \cite{adam:2015}, and
the learning rates in both settings are tuned separately.
In addition, we feed 100k data to the model in one epoch. 
In LB, each batch costs $0.22$s and each epoch costs $330$s. 
In SB, each batch costs $0.13$s and each epoch costs $2285$s. 
  
\begin{figure*}[!t]
	\centering\includegraphics[width=5.0in]{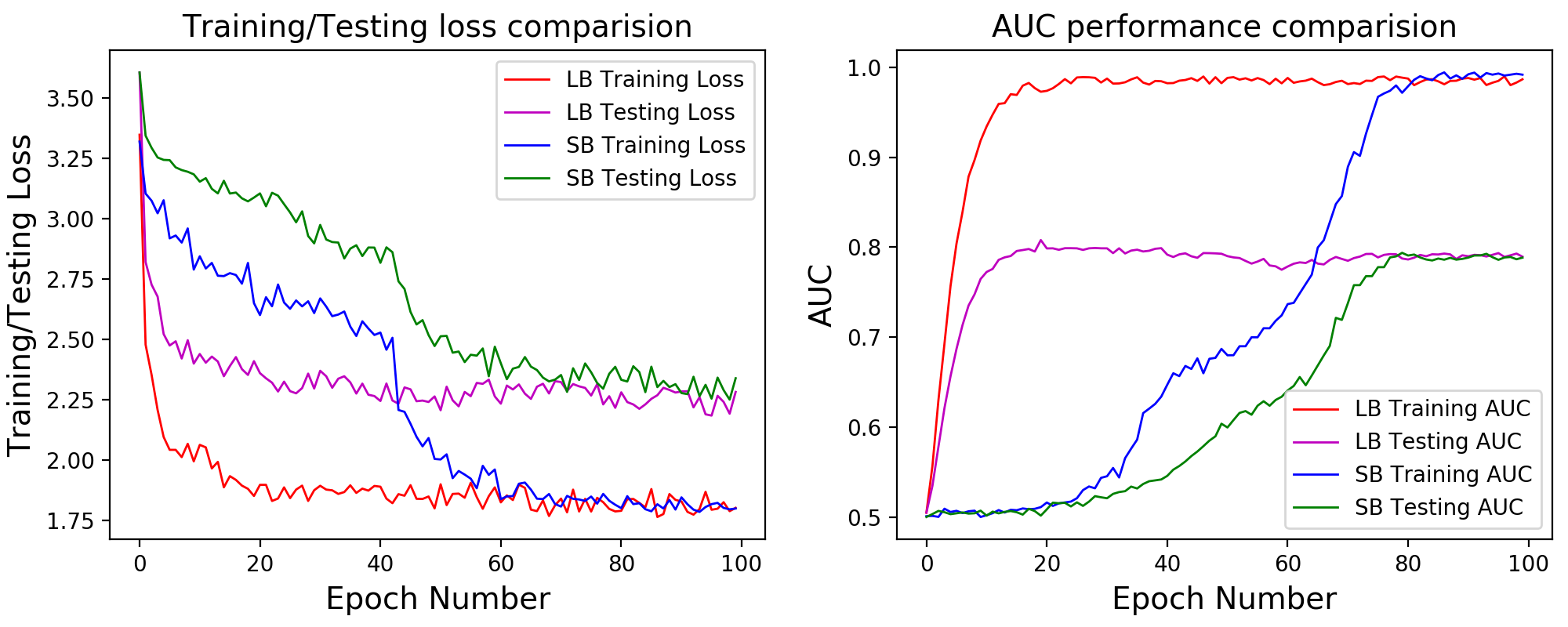}
	\caption{
	The loss and AUC during the training when using LB and SB.
	}
	\label{fig:performance_comparison}
\end{figure*}

We show the loss and the AUC during the training in 
% Table \ref{table:performance_comparison} and 
Figure \ref{fig:performance_comparison}. 
We can see that the optimization for LB is faster (with only 20 epochs to converge) and results in higher AUC on testing data, which indicates that the model trained using LB performances better than using SB.

% The training loss and the AUC on the training data and testing data of LB and SB are shown in Table \ref{table:performance_comparison}. In LB, the loss is minimized to $1.8$ in the $20$th epoch and keeps stable from the $20$th to the $80$th epoch. In SB, the loss is just optimized to $2.6$ in the $20$th epoch and to $1.8$ in the $80$th epoch. LB needs $1.83$ hours to converge and SB needs $50.8$ hours. Besides, the AUC of SB on the testing set in the $80$th epoch is $0.79$, which is less than the best AUC in LB on the testing set. Thus, the generalization of the model with LB is better than SB in our experiment. 

% \subsection{Real World Experiments (Q4)}
\subsection{Online experiments (Q5)}
With the encouraging results on the three real datasets, we perform online experiments on DuerOS.
We deploy this model to the DM module of DuerOS to control whether to ask a clarification question or directly give the candidate response to the user.
Before presenting the results, we first introduce
the experiment setup for the online experiments.

% Having obtained encouraging results on three real datasets, we finally perform real-world experiments on
% a real large-scale spoken dialogue system (DuerOS). 
% We deploy this model on the DM module of DuerOS, aiming to control whether to ask a question or directly give the candidate response for user queries. Before presenting quantitative results, we first detail
% the experimental setup and evaluation metric of our experiments.
\textbf{Experiment setup and evaluation metrics.} 
We use A/B testing, which is originated from \citet{deininger1960human} and
% A/B testing is 
widely used to perform controlled experiments with two or more variants in real-world systems. 
In our experiments, we partition the users into 70/10/10/10 groups for four variants:
In the first group, there is no user satisfaction predictor, i.e., the system directly provides the candidate response to the user. 
In the second group, we use Trans-Text as the user satisfaction predictor. 
In the third group, we use TDU as the user satisfaction predictor.
In the fourth group, we adopt the transformer-based model as the user satisfaction predictor. 
In addition, we train each model with the training samples in the last week and update them every day. All experiments lasted for two weeks to avoid daily fluctuations.
% The experiments last for two weeks to eliminate the daily difference.

We measure the online performance for each model using the contextual user satisfaction (CUS) where the label comes from expert annotators (cf. Section \ref{Evaluation}). To estimate CUS, we first sample the same number of user queries (1000 queries) from the queries in each group.
Then, we extract the context for each query (e.g., 10 turns before the query, 10 turns after the query, and the candidate response), manually annotate whether it is a satisfied query for each sample and calculate the CUS for the target turn. 

% In the evaluation, we first extract the same number of user queries (1000 queries) from each group. Then we extract context around each user query (e.g. ten turns before the current user query, ten turns after the user query, and the information of candidate response). At last, we use the context of each user query to measure whether the candidate response would satisfy the user, whether the system needs to propose the question, and if proposes a question and the user answers this question, whether the final response can satisfy the user. Under such context, if the system directly gave the candidate response for a user query and this response is evaluated to satisfy the user or the system proposed a question and the final response is evaluated to satisfy the user, we consider this query has been satisfied by the system. The final degree of satisfaction (FDS) is defined as the ratio of the number of satisfied users queries to the total number of user queries.

\textbf{Results.} 
The average CUS ratings across samples for the four groups (without a user satisfaction predictor, Trans-Text, TDU and \textbf{the transformer-based model}) are 0.654, 0.662, 0.668 and \textbf{0.684} respectively. 
We see that the transformer-based model achieved the best performance gain in terms of CUS, which is consistent with previous offline experiments.
% Under such evaluation metric, the FDS of the no DM predictor is 66.4\%, the FDS of the Trans-Text is 67.2\% and the FDS of the two-layer transformer is \textbf{68.1\%}. 
% Consistent with the previous discoveries, two-layer transformer method achieved its best performance gain in FDS. 
Given these
promising results, our model has been successfully deployed to the DM module of DuesOs, serving hundreds of millions of user quires on a daily basis.
%引入文献不要反复引用

\section{Conclusion}
In this paper, we propose a transformer based model to predict the user satisfaction for proactive interaction mechanism in a large-scale dialogue system (DuerOS), where we design a transformer-based model to predict the user satisfaction that helps the system decide whether to ask a user a clarification question. Specifically, we generate a large number of weak labels according to the user's interactions with the system in the current turn and the next turn. 
Based on these weak labels, we propose a transformer-based model to extract information from both the structured and text data and grasp the temporal dependency between the current turn and the previous turns for prediction. 
We also find using a large batch size is empirically more effective in the training when the data is noisy.
Furthermore, we conduct experiments on three large datasets and deploy the model to a large-scale spoken dialogue system. We evaluate the proactive interaction mechanism by a new metric (contextual user satisfaction), which can measure the user experience when the system asks a clarification question to the user. The result indicates the effectiveness of our method.

\section{Acknowledgement}
Dr Xuyun Zhang is the recipient of an ARC DECRA (project No. DE210101458) funded by the Australian Government.
% Additionally, we did an experiment to analyze the memorization and the generalization of the model. 

%In the future, we will add the context information of the conversations and the user profile to the features, which may further improve the performance. 
%We will also adopt other noise reduction techniques to reduce 
% the bad impact of the noise. 
%the adverse impacts of the noise from weak labels.

%Based on these weak labels, we proposed a transformer-based model with LB to combine the structured data and text data for prediction. 

%Then, we found that, empirically, large batch size is more effective to optimize the model when the training data is noisy. 

%In addition, we did a series of case studies to analyze the memorization and generalization of the model. 
%which is used to describe the user preference to a given specific result. 


\bibliographystyle{ACM-Reference-Format}
\balance
\bibliography{acmart}

\end{document}